\documentclass[final]{cvpr}

\usepackage{times}
\usepackage{epsfig}
\usepackage{graphicx}
\usepackage{amsmath}
\usepackage{amssymb}
\usepackage{caption} 
\usepackage[pagebackref=true,breaklinks=true,colorlinks,bookmarks=false]{hyperref}
\newcommand{\specialcell}[2][c]{%
  \begin{tabular}[#1]{@{}c@{}}#2\end{tabular}}
  
\newcommand{\beginsupplement}{%
        \setcounter{table}{0}
        \renewcommand{\thetable}{S\arabic{table}}%
        \setcounter{figure}{0}
        \renewcommand{\thefigure}{S\arabic{figure}}%
     }
\begin{document}

\title{A Deep Learning Approach to Private Data Sharing of Medical Images Using Conditional GANs}
\author{ 
    \specialcell{Hanxi Sun\thanks{Co-first authors}\\
    Purdue University, Department of Statistics\\
    West Lafayette, IN, USA}
    \hspace{1cm}
    \specialcell{Jason Plawinski\textsuperscript{\textasteriskcentered}\\
    Novartis\\
    Basel, Switzerland}
    \and
    \specialcell{Sajanth Subramaniam\\
    Novartis\\
    Basel, Switzerland}
    \hspace{1cm}
    \specialcell{Amir Jamaludin\\
    Oxford Big Data Institute\\
    Oxford, UK}
    \hspace{1cm}
    \specialcell{Timor Kadir\\
    Oxford Big Data Institute\\
    Oxford, UK}
    \and
    \specialcell{ Aimee Readie\\
    Novartis\\
    East Hanover, NJ, USA}
    \hspace{1cm}
    \specialcell{Gregory Ligozio\\
    Novartis\\
    East Hanover, NJ, USA}
    \hspace{1cm}
    \specialcell{
    David Ohlssen\\
    Novartis\\
    East Hanover, NJ, USA}
    \and
    \specialcell{Mark Baillie\thanks{Co-last authors}\\
    Novartis\\
    Basel, Switzerland}
    \hspace{1cm}
    \specialcell{Thibaud Coroller\textsuperscript{\textdagger}\thanks{Corresponding author: {\tt thibaud.coroller@novartis.com}}\\
    Novartis\\
    East Hanover, NJ, USA\\
    }
}
\maketitle

\begin{abstract}
Sharing data from clinical studies can facilitate innovative data-driven research and ultimately lead to better public health. However, sharing biomedical data can put sensitive personal information at risk. This is usually solved by anonymization, which is a slow and expensive process. An alternative to anonymization is sharing a synthetic dataset that bears a behaviour similar to the real data but preserves privacy. As part of the collaboration between Novartis and the Oxford Big Data Institute, we generate a synthetic dataset based on COSENTYX\textsuperscript{\textregistered} (secukinumab) Ankylosing Spondylitis clinical study. We apply an Auxiliary Classifier GAN to generate synthetic MRIs of vertebral units. The images are conditioned on the VU location (cervical, thoracic and lumbar). In this paper, we present a method for generating a synthetic dataset and conduct an in-depth analysis on its properties along three key metrics: image fidelity, sample diversity and dataset privacy.
\end{abstract}

\section{Introduction}
In recent years, deep learning has become an indispensable tool for clinical image analysis ranging from ophthalmology\cite{fauw}, pathology\cite{schmauch} to medical imaging\cite{hosny, xu}. By the end of 2020, PubMed\cite{pubmed} had a total of 9,497 entries for “deep learning + image” with more than half of them published in 2020. While the number of publications using such techniques has exploded, the number of publicly available datasets has not. Many institutions are still unable to share data due to privacy concerns. It results in each research group working independently on its own databases with no pooling and centralization of the data. This drastically reduces the impact of each individual analysis. The lack of data availability also impacts reproducibility and replication studies\cite{naos}. Most method papers are not published with the corresponding research data, which restrains other scientists from verifying results and validating hypotheses. More generally, sharing medical data is arduous due to privacy\cite{smidt, tom}, ethical\cite{wing}, legal\cite{cannataci}, and institutional challenges\cite{grama}.

Despite some notable efforts in data sharing\cite{prior}, privacy remains a major hurdle in the democratization of datasets. The most common way to share data while preserving privacy is by anonymizing it. One way of doing so is by applying risk-based anonymization\cite{emam}. Risk-based anonymization aims at controlling the likelihood  that a patient can be re-identified by using the available data. Though risk-based anonymization can allow fine privacy control, it often relies on de-identification or on dropping features that contain sensitive information.

Another type of anonymization technique is differential privacy\cite{dwork} where carefully crafted noise is added to samples during a query process. This solution is one of the leading anonymization methods on tabular data because it offers strong, customizable, mathematically grounded privacy guarantees. However, differential privacy does not translate well to the image domain. Adding noise to individual pixels only fundamentally alters the image on a mathematical level while preserving higher resolution features and image content, this makes the effectiveness of any pixel-based anonymization elusive. Efficient implementations of differential privacy rely on altering data during each new query process, this means that data needs to be hosted on internal servers so it can be altered before being shared. Because the standards for image datasets do not rely on query-based databases, it is practically much more complicated to apply differential privacy to image data. There is little related work available on the direct applications of differential privacy to images with most work still being limited to feasibility studies\cite{fan}. Literature on privacy application to images is focused more on making the training process of neural networks differentially private\cite{abadi} than transforming sensitive data into differentially private data.



Federated learning\cite{ryffel} is another solution to privacy challenges: in the federated learning framework, instead of sharing the data, it is the model that is being shared between scientists. Each team then trains the model on their own private data rather than by pooling all the datasets together. This solution totally bypasses the need for anonymization but the trade-off is that the model itself needs to guarantee privacy. More precisely, the training samples should not be identifiable by reverse engineering the trained model. An additional limitation is that all the researchers have to work with one single model and are unable to prototype on the other teams' private datasets.

Responsible data sharing can be tackled by sharing a synthetic version of a sensitive dataset that preserves global properties while preserving patients privacy. While nothing will ever replace an actual dataset, an approximate version can help for important research steps. For example, a synthetic dataset could be used to prototype new methods and new architectures, be used for data challenges and hackathons or as surrogate data for research papers. The privacy aspect of synthetic dataset is closely related to risk-based anonymisation: unlike differential privacy which upholds strict privacy guarantees by design, the privacy assessment of synthetic datasets has to be done by evaluating the risk of re-identification.

\begin{figure*}
\begin{center}
\includegraphics[width=\textwidth]{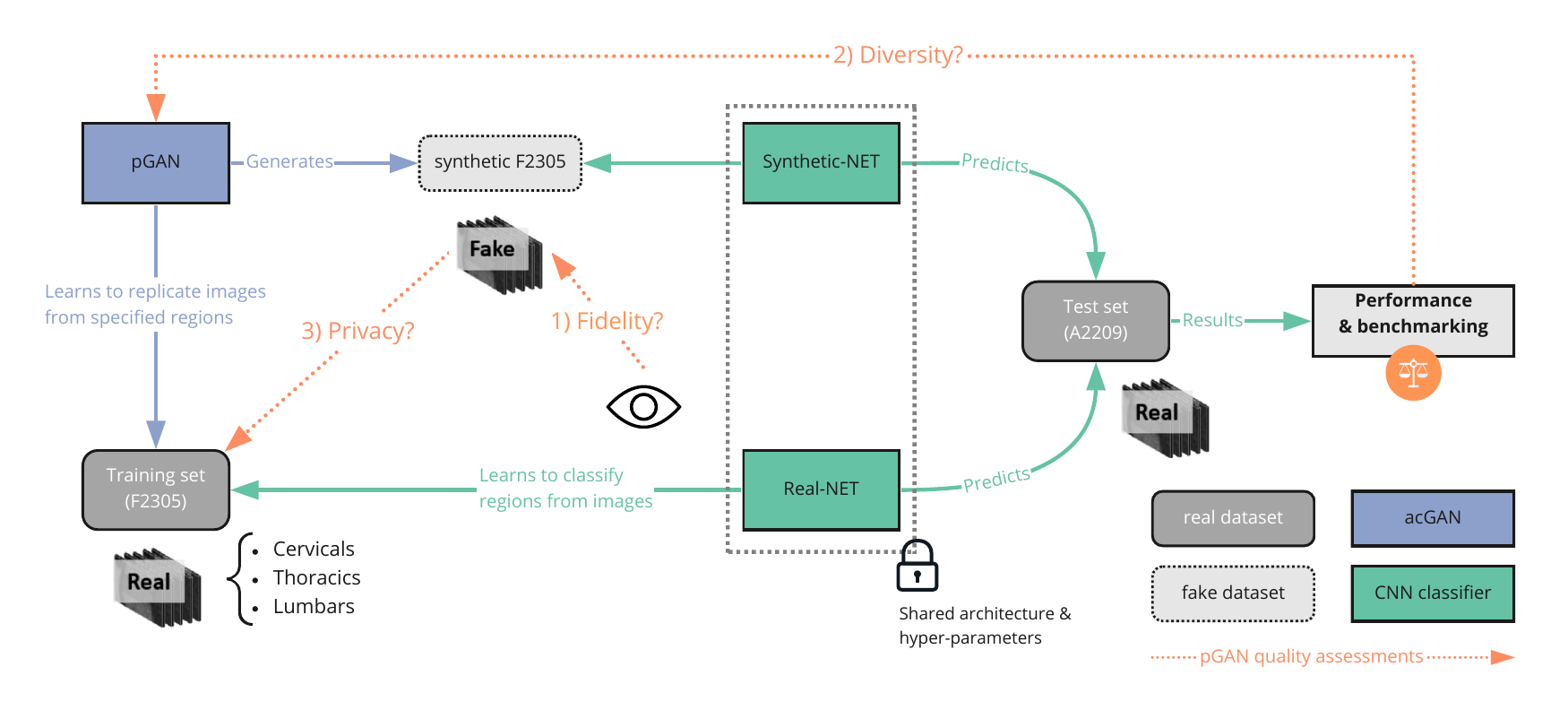}
\end{center}
   \caption{Overview of the privacy computing workflow to share synthetic medical images (vertebra MRIs) with our privacy GAN model (pGAN). We trained an auxiliary classifier generative adversarial network (AC-GAN) from a dataset composed of vertebras images alongside their corresponding locations (cervical, thoracic, lumbars). We then used our trained pGAN to generate a synthetic dataset of vertebrae. The quality of the output synthetic dataset was evaluated based on three criteria (privacy, fidelity and diversity). With all criteria passed, we can safely share those synthetic data with external scientists without any privacy risk while retaining most of the data usefulness.}
\label{fig:01}
\end{figure*}

In this paper we propose to use generative adversarial networks (GANs)\cite{goodfellow} to create synthetic images while conserving their association with clinical variables (as shown in Figure \ref{fig:01}). We define three core properties that a synthetic dataset should display to be shared externally as a meaningful substitute of the original dataset: fidelity, diversity and privacy. We then confirm that our synthetic dataset produces realistic samples (fidelity), that the samples exhibit similar variations as the original dataset (diversity), and that no original sample can be retrieved from the synthetic dataset (privacy).

\section{Results}
To evaluate the effectiveness of the GANs as a privacy method, we create a synthetic dataset of T1 magnetic resonance images (MRIs) of vertebral units (VUs) labelled with VU location (cervical, thoracic and lumbar) and assess the fidelity, diversity and privacy of the synthetic dataset.

\subsection{Fidelity}
\begin{figure*}
\begin{center}
\includegraphics[width=0.9\textwidth]{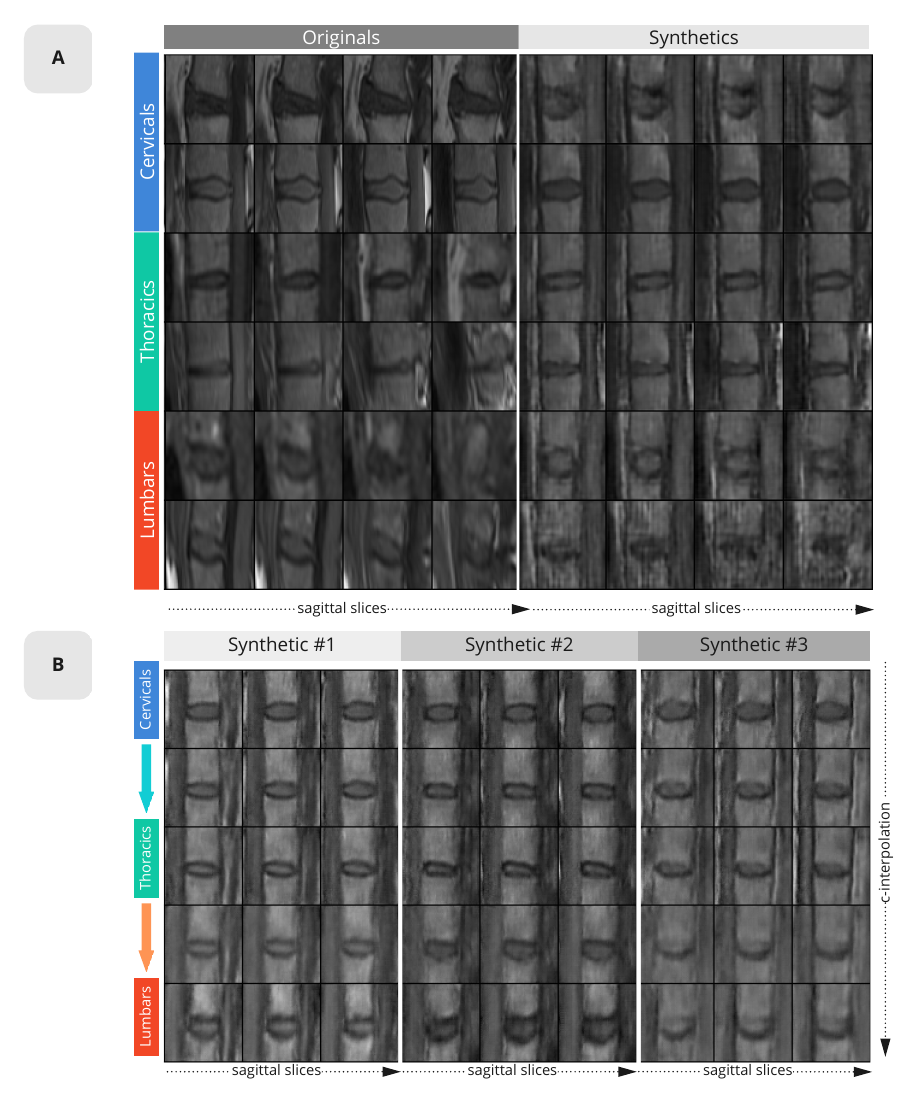}
\end{center}
   \caption{A) Examples of real and synthetic vertebrae and their corresponding vertebrae units (VU) locations. Each row shows 4 consecutive vertical slices for a real and fake VU (each slice is of size 64x64). The first two rows are cervical VUs, followed respectively by thoracic and lumbar. B) Morphing an image from one location to another was done by interpolating the condition $c$ while keeping the latent variable $z$ untouched. Here we are using a single step $n=1$ for each morphing such that the shown intermediate states correspond to a 50\%/50\% morph between two locations.}
\label{fig:02}
\end{figure*}

We start by assessing the fidelity of the generated images by evaluating whether the synthetic VUs are visually similar to real VUs for a given location. Figure \ref{fig:02}-A shows a collection of real and synthetic data. Assessing perceptual quality of images in a quantitative manner is an open topic with which the computer vision community has been struggling for decades. The most successful perceptual metrics used to grade synthetic images are Inception Score (IS) and Frechet Inception Distance (FID). Both metrics work by comparing the distribution of low or mid level image features from real and generated images. These techniques therefore require a pretrained network. Unfortunately, no such model is available for our data type (3D MRI with 9 channels).

We therefore had to rely on visual inspection to assess the fidelity of the GAN model. Synthetic samples look similar to real samples and display region specific features depending on the conditioned location (i.e. cervical, thoracic and lumbar). Inspired by the works on face morphing with GANs\cite{karras}, we show in Figure \ref{fig:02}-B that the generation behaviour can be controlled by continuously changing the condition vector $c$. This is done by linearly interpolating between the source and target region wanted. The fact that interpolated VUs look realistic strengthen the idea that realistic, high fidelity samples can be generated for any combination of input parameters.

\subsection{Diversity}
\begin{figure*}
\begin{center}
\includegraphics[width=0.85\textwidth]{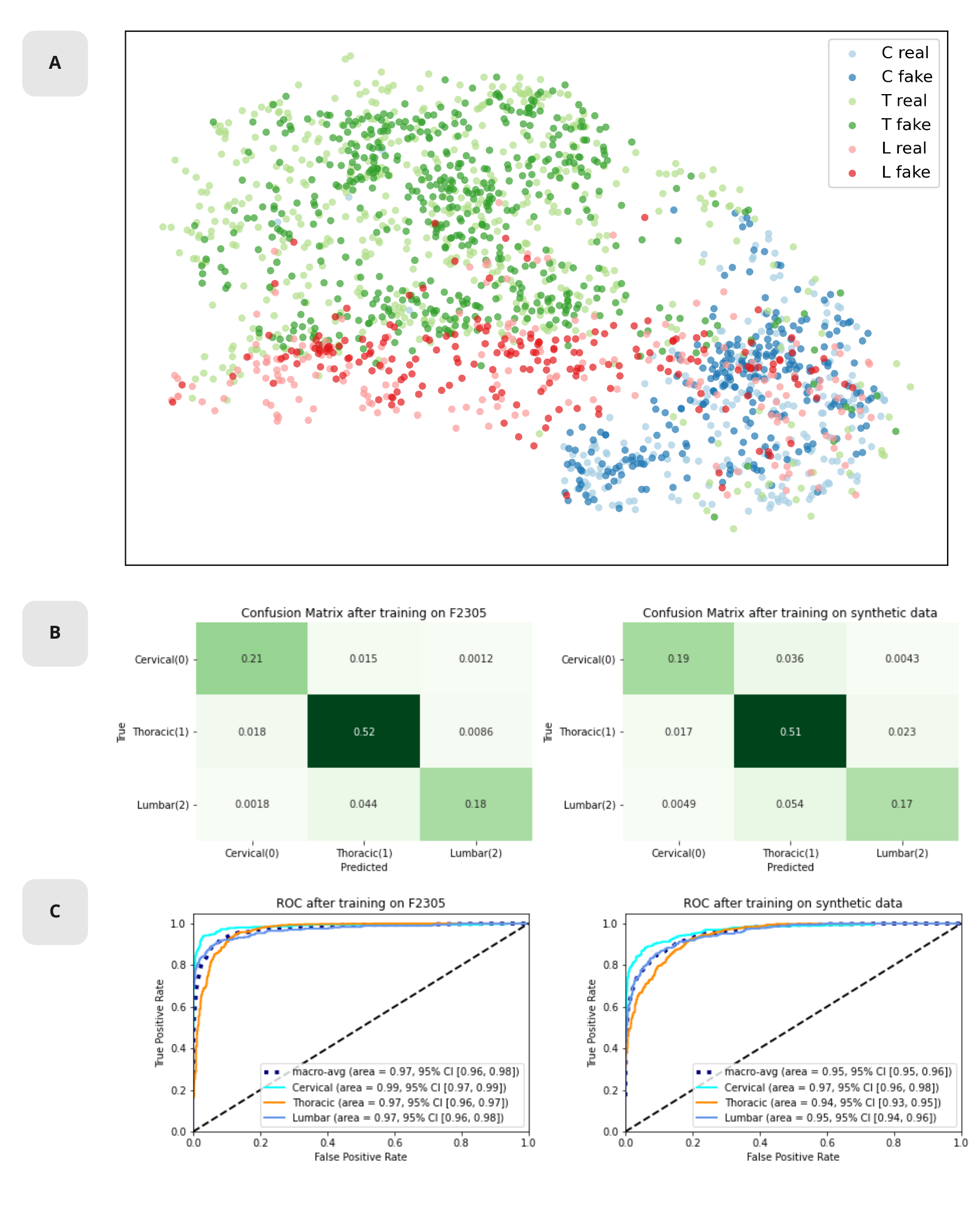}
\end{center}
   \caption{A) Low dimensional UMAP representation of real and synthetic vertebra units locations cervical, thoracic and lumbar labelled as blue, green and red respectively. B) Confusion matrices of $F_\text{real}$ (left) and $F_\text{synth}$ (right) when tested on A2209. The predicted class for a sample image $x_i$ corresponds to $\max(F_\text{real}(x_i))$ and $\max(F_\text{synth}(x_i))$ respectively. The class distribution of the test set corresponds to $(p_C,p_T,p_L)=(0.23,0.55,0.22)$. C) ROC and AUC for $F_\text{real}$ (left) and $F_\text{synth}$ (right) when tested on A2209. Every class specific ROC curve was computed by treating the other two classes as negative instances, effectively reformulating the classification for each class as a binary problem.}
\label{fig:03}
\end{figure*}

In this section we evaluate the diversity of the synthetic dataset. This analysis is conducted at a dataset level (instead of at a sample level like for fidelity).
We compare whether the real and synthetic dataset have similar global distributions as seen on Figure \ref{fig:03}-A. To obtain a meaningful low dimensional representation of the synthetic and original dataset, we apply a UMAP\cite{mcinnes} transform to our datasets of VUs. UMAP is a powerful non-linear dimensionality reduction technique. It is similar under many aspects to t-SNE\cite{vandermaaten} but one key difference is that the transformation from high dimension to low dimension is learnable. We leverage this by training the UMAP dimensionality reduction exclusively on real data and then applying the learned transform to unseen real data as well as synthetic data. This is useful because the transformation is not biased by synthetic samples and only represents significant features from the real dataset.

We can observe from Figure \ref{fig:03}-A that the synthetic samples span the same global region as the real samples. The original dataset also appears separable based on VU location in the UMAP space and this behavior is well preserved during synthetic generation. The general distribution of real and synthetic data behaves in a similar manner. However, only few synthetic samples lay on the edge of the distribution meaning the GAN generates a more conservative dataset with less fringe and rare cases.

We further examine the quality of the synthetic dataset in terms of its capability to serve as a functional equivalent to a real dataset for downstream training tasks by training two ResNet-18\cite{he} based classifiers $F_\text{real}$ and $F_\text{synth}$ with real and synthetic data respectively. Figure \ref{fig:03}-C shows the ROC curves for both classifiers. We observe that, while $F_\text{synth}$ slightly underperforms compared to $F_\text{real}$, the synthetic dataset can approximate the distribution in the original data when tested for VU region. The slight discrepancy in performance is unsurprising as the conservation of privacy can be expected to come at a cost in data quality. While, in principle, the performance of $F_\text{synth}$ could asymptotically reach the one of $F_\text{real}$, in practice there is a trade-off between overfitting the training set and replicating the classification statistics of a network trained solely on the training set.

\subsection{Privacy}
We define a privacy leakage as finding traces of the training dataset within the synthetic dataset. In this section, we restrict the study to only sharing a synthetic and finite dataset rather than the trained generative model itself. This restriction makes privacy attacks much more difficult because they cannot be conducted by reverse engineering the network’s output and there is no direct access to the generative model’s latent space. This means only attacks that rely on directly comparing candidate images with the synthetic dataset can jeopardise privacy. We evaluate privacy in a risk-based manner by measuring the likelihood of re-identification of patients from the training set.

We identified two main potential vulnerabilities for a synthetic dataset:
\begin{itemize}
    \item \textbf{Pairwise attacks}: where finding a synthetic image which is similar to a given sample would prove that this sample was used during training.
    \item \textbf{Distribution attacks}: where a high density of synthetic images cluster around one or a few real images.
\end{itemize}

\begin{figure*}
\begin{center}
\includegraphics[width=\textwidth]{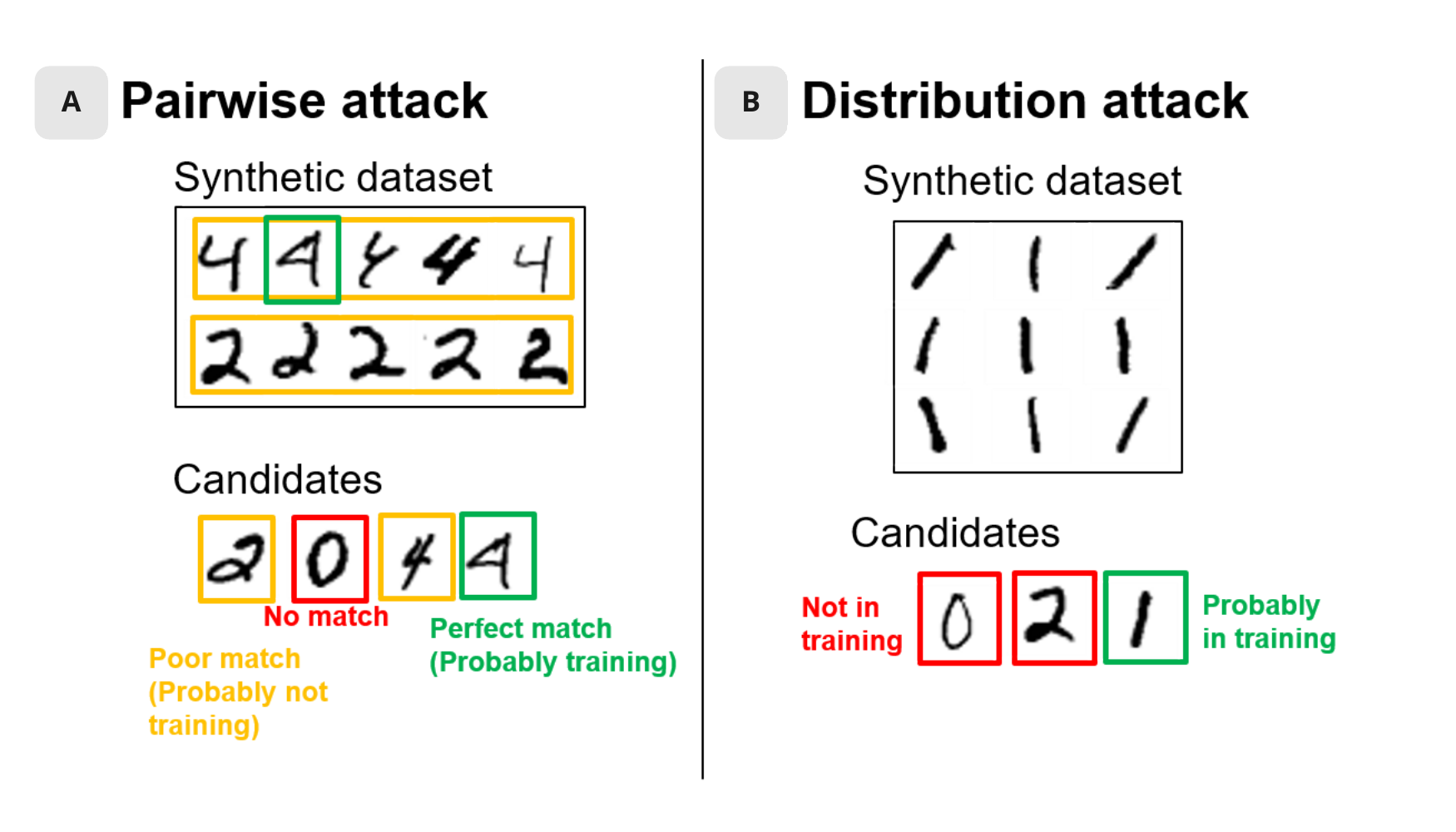}
\end{center}
   \caption{\textbf{Candidate pair definition.} Column A: example of a pairwise attack of a synthetic dataset. The green digit from candidates closely matches the green digit from the synthetic dataset meaning this candidate is likely in the train set and privacy is not preserved. Column B: example of a distribution attack on a synthetic dataset. While not being an exact match, the green digit is very similar to many digits from the synthetic dataset. This indicates that the green digit is likely present in the train set.}
\label{fig:s01}
\end{figure*}

To assess the robustness to these attacks, we simulate them (supplementary Figure \ref{fig:s01}) by using a “candidate dataset” composed in equal proportion by training, validation and test samples. We then evaluate which samples can accurately be traced back to the training set.

\subsubsection{Robustness to pairwise attacks}
\begin{figure*}
\begin{center}
\includegraphics[width=\textwidth]{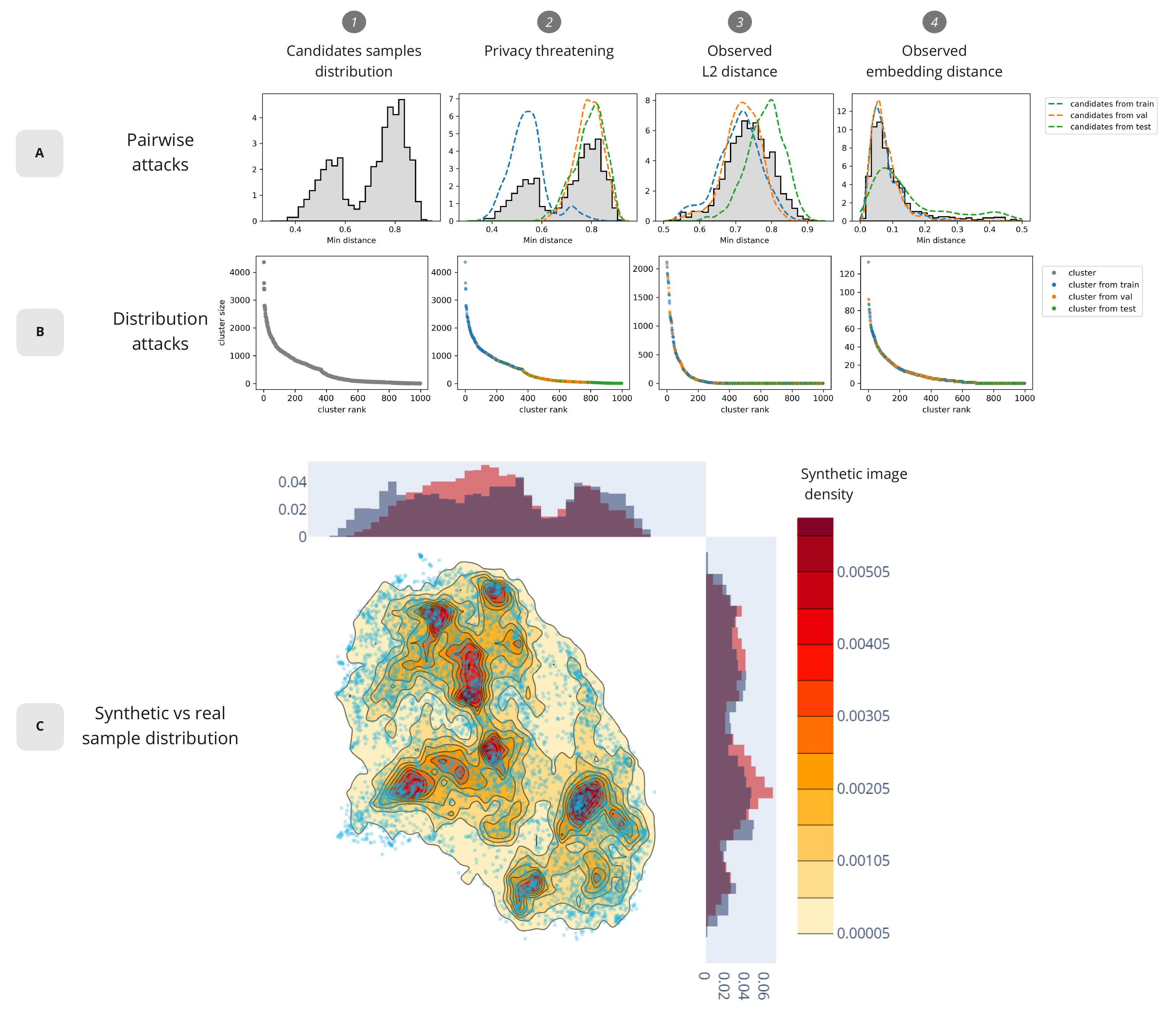}
\end{center}
   \caption{A) \textbf{Simulation of pairwise attacks.} A small distance indicates that the candidate is likely from the training set. 1-2 are toy examples of a privacy threatening scenario with easily identifiable candidates from training. 3-4 are the observed behavior of our synthetic dataset. B) \textbf{Simulation of distribution attacks.} Large cluster indicates that the candidate is likely coming from the training set. 1-2 are privacy threatening and 3-4 are observed behavior on the synthetic dataset. Cluster size refers to the number of neighbours of a candidate sample and a cluster rank n means the cluster is the n\textsuperscript{th} biggest cluster of the set.  C) 2D projection of the synthetic and training dataset. Shades of yellow-red are density levels of synthetic images and blue dots are individual training examples.}
\label{fig:04}
\end{figure*}

\begin{figure*}
\begin{center}
\includegraphics[width=\textwidth]{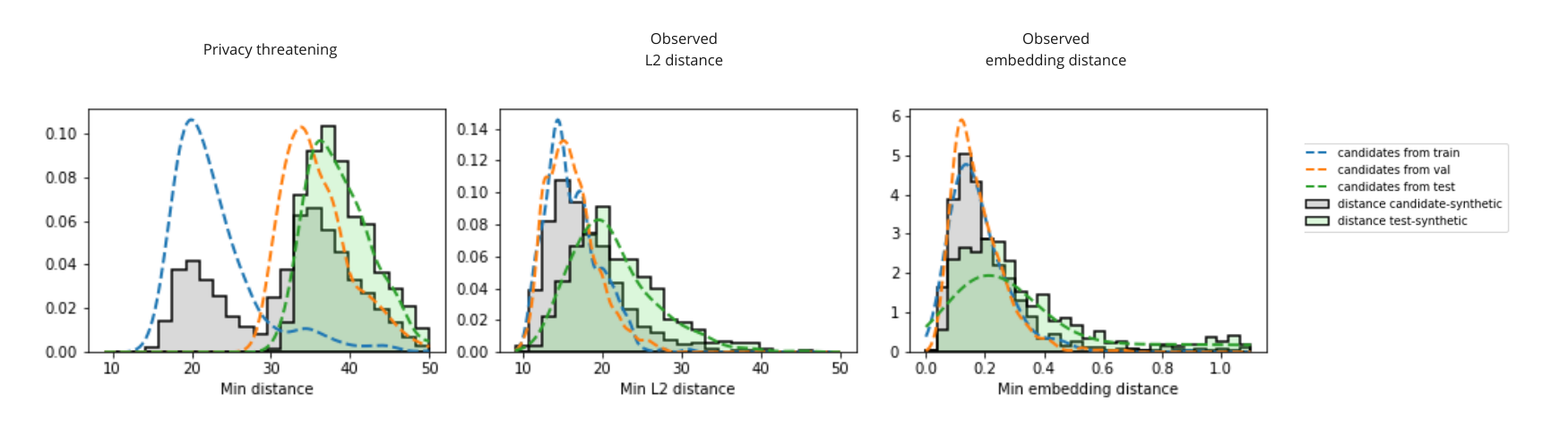}
\end{center}
   \caption{\textbf{Pairwise attack with reference.} In this experiment, the abnormally low distances between candidates and synthetic can be identified by anomaly detection. An abnormal sample is any sample that diverges largely from the distance test-synthetic (green distribution). For our synthetic dataset, validation samples are also classified as outliers, privacy is preserved.}
\label{fig:s02}
\end{figure*}

For robustness to pairwise attacks, we compute the similarity between a candidate sample and all the images from the synthetic dataset. A relatively small L2 distance may indicate that the candidate was used for the generator training. The distribution of minimum distances between candidate and synthetic images is represented in Figure \ref{fig:04}-A. On this graph, we would expect the lowest distances to be train-synthetic pairs and the largest ones to be test-synthetic due to some overfitting behaviour from the GANs. In the privacy-threatening scenario, images from training are easy to identify by a simple anomaly detection task. An anomaly detection with reference distances is available on supplementary Figure \ref{fig:s02}. On our observed synthetic dataset however, it is impossible to reliably identify candidates originating from training from the ones coming from validation. This is also confirmed using an embedding space obtained by applying a UMAP. Our synthetic dataset hence shows robustness to pairwise attacks, in both pixel and embedding space.

\subsubsection{Robustness to distribution attacks}
To evaluate the robustness to distribution attacks, we identify clusters of synthetic images around a given candidate. The clusters are defined by counting the number of synthetic images in the candidate neighbourhood and the results are shown in Figure \ref{fig:04}-B. We expect large clusters of synthetic data to form around candidates from the training set and other candidates to have a small number of synthetic neighbours. This is reflected on the privacy-threatening scenario on the left figures from \ref{fig:04}-B. The observed behaviour of the synthetic dataset is shown on the two most right figures. Candidates from the test set rarely belong to large clusters unlike candidates from training or validation. Even if this method can help identify a handful of training examples it stays limited by the size of the synthetic dataset. Indeed, the size of the training set and the synthetic set are around the same (around 10’000 samples each). This means that by construction, our synthetic dataset is safe from distribution attacks because clusters cannot be meaningfully large without first heavily loosening the definition of neighbourhood.

It is also worth noting that the embedding space is not uniformly populated. As seen on Figure \ref{fig:04}-C, high density of synthetic images is in general linked to a high density of real images (blue dots on the figure) and not due to the GAN generation collapsing. A detailed detection of training samples is available in the supplementary material Figure \ref{fig:s03}, Figure \ref{fig:s04} and Table \ref{tab:label}.

\section{Discussion}

Sharing data across research groups and institutions provides an opportunity to answer complex questions through the pooling of information and resources. However, there is a need to share data safely and faster. Sharing synthetic dataset overcomes the privacy and legal barriers to enable efficient collaborations. We propose generating synthetic datasets of labelled medical images with a type of GAN model. It is worth noting that besides privacy protection, GANs have been used often in medical research, mostly as a data augmentation\cite{kazeminia, yi} approach, see for instance \cite{chen, shin, han, fossen-romsaas, frid-adar, zhang}. To create a synthetic dataset we used an AC-GAN to generate synthetic samples with their respective location labels. The synthetic dataset can safely be shared as a surrogate for the real dataset because it meets the consistency and privacy criteria.
\begin{itemize}
    \item \textbf{Fidelity:} The synthetic samples are realistic and bear high visual quality of samples.
    \item \textbf{Diversity:} The behaviour of the real dataset was replicated and dataset wide analysis such as classification (trained from scratch) performed to a similar level on both the synthetic and the real dataset.
    \item \textbf{Privacy:} No sample from the synthetic dataset can be associated with a sample from the real dataset making it impossible to trace back patient information from a synthetic sample. 
\end{itemize}

In our case, we decided to only share a fixed sized dataset to avoid privacy concerns coming from the model itself. The major privacy attack we have to be wary of are membership inference attacks\cite{mukherjee}. They aim at determining whether data from a target patient is included in the study. A general approach to protect data from such attacks is differential privacy\cite{dwork}, which masks real data with a carefully designed noise during training. It provides a strong protection of privacy and has been applied in a variety of problems, including synthetic data generation with GAN\cite{jordon, chen, beaulieu}. We do not compare our results against pure differential privacy because the effectiveness of noise-based methods on images are hard to evaluate. On images, noise only alters mathematical similarities while keeping general shapes and image content intact. For example the effectiveness of differential privacy (like additive Laplacian noise) can, to some extent, be mitigated by CNN denoisers\cite{fan} which makes it an imperfect benchmark.

We only evaluate privacy through the re-identification of training samples (by outlier detection) which, in our case, is the only potential type of attack. Our results show our GAN did not replicate samples from the training set and tends to be robust to pairwise attacks. The presence of artifacts, for example, affects the similarity measures in pixel space rendering the comparison of real and synthetic samples particularly difficult. Comparing in feature space does not solve the problem because the only embedding space where synthetic and training images share similar features is the GAN embedding space itself. Density attacks seem to be more dangerous than pairwise attacks because it does not require an exact match between a synthetic and a candidate sample. Simply having multiple synthetic images that are like a candidate image could already be enough to conclude on its origin. Thankfully, by sharing a dataset of small enough size our synthetic dataset is robust to density attacks. With around 1.2 synthetic data points per real data point, our synthetic dataset is too small to create meaningful clusters. This means clusters will either be large but composed of barely similar images or too small to draw conclusions.

The quantitative analysis available in the supplementary material shows the limited discriminative power of outliers detection to identify training samples. On the table \ref{tab:label}, we can see that by selecting the top 50 outliers identified either by smallest distance or by largest clusters, at best two thirds of the outliers are rightfully attributed to the training set while one third in fact belong to the validation set. When selecting all probable training samples as outliers (top 333 out of 1000 candidate samples), the validation and training samples are nearly indistinguishable. We can conclude that, for these specific challenging conditions, the privacy of the dataset is well preserved. However, these guarantees might not be enough for all applications and privacy is not guaranteed to hold for some more extreme cases. Defining acceptable privacy thresholds based on the sensitivity of the data and the downstream application seems an important direction for further investigation. This is especially crucial in our case because privacy is not tied to a single, easily tweakable hyper parameter like with $\epsilon$ for differential privacy.

\section{Limitations}
Ensuring privacy is a challenging aspect of this project, it is particularly difficult to define a robust method to assess similarity between images\cite{wang}. Due to the limited amount of samples that were available, we only investigated generating VUs instead of full spine images. For computing efficiency, we resampled the images to a smaller format. We also tried conditioning the AC-GAN on clinical metrics but the imbalance between the number of positive samples and negative samples as well as the intra-reader discrepancies led to similarly noisy results that proved too hard to analyse. Finally, in our approach, the synthetic dataset has a fixed privacy tolerance. This fixed tolerance is a factor of the real dataset and the GAN's convergence and cannot be modulated based on specific needs.

\section{Future steps}
Even though the synthetic dataset produced in this evaluation does not contain synthetic samples close to real samples, the generator can theoretically create such images. It is unlikely to happen as the curse of dimensionality greatly reduces the probability of sampling an image with the exact same parameters as the real data. However, having to check that the network did not reproduce a real sample after each generation is inefficient. A direction for further research could be to learn where the real images lie in the latent space of the generator and then create a careful sampling strategy to avoid sampling around these points. Another research direction that is considered is to condition the generation process not only on a few classes but on large tabular data (such as clinical and demographic data). In fact, the current architecture would have to change significantly. Another step of privacy preserving synthetic generation, this time of tabular data would be necessary. Additionally, the loss function of the AC-GAN does not behave well with soft labels and conditions with small or no correlation with the content of the images. On the spine dataset, a useful feature would also be to generate a full spine and instead of independent VUs. Finding a good trade-off between VUs consistency and memory is important however since generating 21 VUs at once is extremely memory intensive while only a weak correlation between VUs would be required. 

\section{Methods}
\subsection{Dataset descriptions}

This work is based on the anomymized dataset of the secukinumab clinical trials for Ankylosing spondylitis. We used MEASURE 1 F2305 (CAIN457F2305) study\cite{baeten}, for which imaging was done using T1 and STIR sagittal MRI at baseline and weeks 16, 52, 104, 156 and 208. For our work, we discarded STIR and pooled all T1 images regardless of their acquisition date. We then extracted 23 vertebral units (VU: section between two adjacent vertebrae) using SpineNet\cite{jamaludin}. F2305 was used for the training and validation, with respectively 7832 and 2205 vertebral units (VUs). The validation set and training set are split so that there is no overlap of patients in both sets. The validation set is exclusively used for the privacy assessment, the GAN hyperparameter tuning was done based on training set metrics. The dataset from the A2209 (CAIN457A2209) study, is composed of 1625 VUs and was solely used for testing purposes.
Each VU was pre-processed to have 9 slices (4 before and after from the central one) and labelled according to its anatomical region (cervical, thoracic and lumbar). VUs from the sacrum were excluded from the pooled dataset. All VUs were rotated to align together vertically and zoomed to have consistent a size-ratio.

\subsection{Auxiliary classifier Generative adversarial networks (AC-GAN)}
\begin{figure*}
\begin{center}
\includegraphics[width=0.9\textwidth]{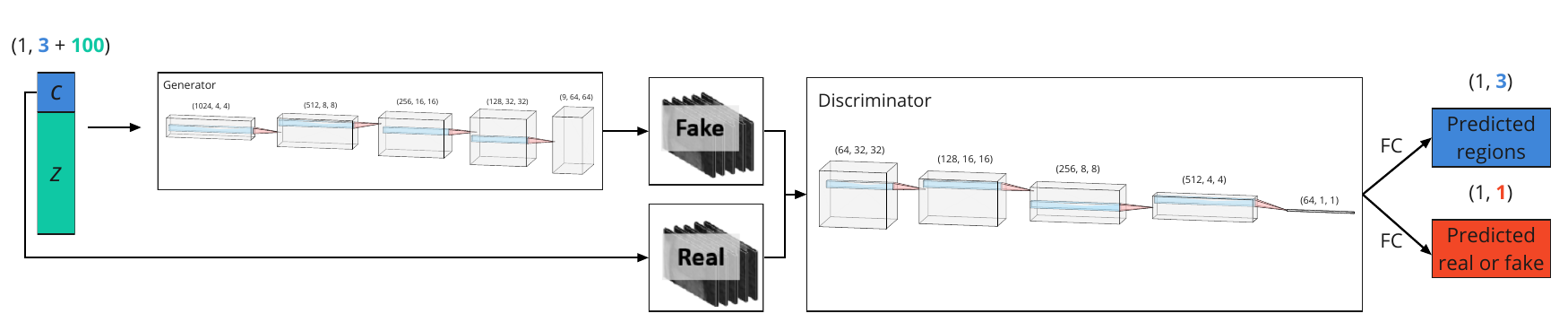}
\end{center}
   \caption{\textbf{The AC-GAN structure.} The generator (G) is a deep neural network that transforms a feature vector of the class label ($c$) and a random noise ($z$) to an image. The discriminator (D) distinguishes between the real images and fake images and also predicts the class labels for any image.}
\label{fig:05}
\end{figure*}

We use auxiliary classifier generative adversarial networks (AC-GAN)\cite{odena} to generate synthetic VUs and associated VU regions. AC-GAN is a variation of generative adversarial networks (GAN) that relies on the competition between two deep neural networks, the generator and discriminator, to create realistic images. Figure \ref{fig:05} shows the general structure of the AC-GAN model.

The training is done by iteratively improving the generator and discriminator. The loss consists of two components, the discriminative loss $L_D$ for distinguishing between real and fake images and an $\alpha$ weighted classification loss $L_C$ for predicting class labels, i.e.
\begin{equation*}
    L = L_D + \alpha L_C
\end{equation*}
where
\begin{gather*}
    L_D = E\left[\log P(\text{real}|X_\text{real})\right]+E\left[\log P(\text{fake}|X_{\text{fake}})\right] \\
    L_C = E\left[\log P (\hat{c}=c|X)\right] 
\end{gather*}

\begin{itemize}
\item \textbf{Pre-processing:} we keep the same 9 slices of VUs and rescale each slice from 112x224 to 64x64. For this specific dataset of VUs, although being 3D images, VUs are treated as 2-dimensional images with 9 channels and learnt with a 2D AC-GAN model, as the 3-dimensional structure in VUs are not very strong due to large slice thicknesses.

\item \textbf{Feature vector:} apart from the additional 3 dimension for class labels (region encoded as a length 3 one-hot vector), we used a 100-dimensional standard Gaussian random noise as the input to the generator.

\item \textbf{ Generator:} the generator consists of five 2-dimensional deconvolutions (transposed convolution) layers with 512, 256, 128, 64 and 9 output channels, respectively. We used a kernel of size 4 in all layers. The first layer has a stride of 1 and 0 padding to upsample the image size from 1x1 to 4x4. The rest layers are set with stride size 2 and padding 1 to double the image size. Rectified Linear Unit (ReLU) activations were used in the first 4 layers and the hyperbolic tangent ($\tanh$) activation was used in the last layer to restrict the output pixels within the range of [-1, 1].

\item \textbf{Discriminator:} the discriminator consists of five 2-dimensional convolution layers with 64, 128, 256, 512, 64 output channels, respectively, and each layer is set with kernel size 4. A stride size 2 and padding 1 is used in all layers but the last one. The last layer is set with stride size 1 and padding 0. Leaky ReLU activations were used in all layers with a negative slope of 0.2. After passing through the convolution layers, VUs are then going through two separate fully connected layers with sigmoid and softmax activation, respectively, to produce the decision between real vs fake as well as the predicted classification label.

\item \textbf{Loss:} the classification loss is weighted by $\alpha=1$ during training.
\item \textbf{Optimization:} Training is done with Adam optimizer\cite{kingma} with learning rate $10^{-4}$ for both the generator and the discriminator.

\item \textbf{Model Selection:} When performing hyperparameter search for training the model, we rely on visual inspection, adversarial and classification loss during training as well as classification on the A2209 test set.

\item \textbf{Synthetic Data:} to generate the synthetic dataset, we first simulate VU location from the empirical distribution of it in the real data. We then use the sampled VU location to condition the generator and generate the entire image.

\item \textbf{Reproducibility:} Trial data used for the GAN training is restricted, but synthetic VU data used for the diversity results are available. Additionally, all Python codes are fully available.
\end{itemize}

\subsection{Real vs. synthetic ResNet-18 performance}

Prior to the training of $F_\text{synth}$, we created a synthetic dataset using our trained generator. We designed the synthetic dataset size and class distribution to approximate F2305 by generating 10,000 synthetic images whereas the class assignment is sampled from a distribution $(p_C,p_T,p_L)=(0.25,0.55,0.2)$ where we denote the probability for cervical, thoracic and lumbar respectively. 

We train two ResNet-18 based classifiers $F$ for the VU region. For the first classifier $F_\text{real}$ we use F2305 as the training set while the second classifier $F_\text{synth}$ is trained solely on synthetic data. In both cases, A2209 was used as independent test set. Both models use the same hyperparameters and an identical training procedure.
\begin{itemize}
    \item Optimization: We use stochastic gradient descent with a learning rate of $10^{-4}$ and momentum of 0.9. Networks were trained for 20 epochs with a batch of 32.
    \item Loss: We use the cross-entropy loss.
\end{itemize}

We compute a single global ROC curve for each classifier by averaging the class specific ROC curves using equal weights (macro-average). The 95 percentile confidence intervals for the AUCs were computed using bias-corrected and accelerated bootstrapping\cite{efron} with 2000 resamples.

\subsection{Image morphing}
To generate our synthetic image, we input a tensor that concatenates a random noise $z \in \mathbb{R}^{100}$ and a condition $c_k \in \mathbb{R}^K$. The tensor $c_k$ is a hot-encoding vector representing the vertebra location
\begin{align*}
    c_{k,i}=\delta_{ki}
\end{align*}
where $\delta$ is the Kronecker delta.
In the case of $K=3$ we have
\begin{align*}
    c_1=[1, 0, 0]^\text{T} && \text{(cervical)}\\
    c_2=[0, 1, 0]^\text{T} && \text{(thoracic)}\\
    c_3=[0, 0, 1]^\text{T} && \text{(lumbar)}
\end{align*}

To morph an image from a location to a new one, we only need to modify our tensor $c$ to impact the generator, thus leaving untouched $z$. We use linear interpolation between source condition $c_k$ and the target one $c_{k'}$. For $n$ steps we have for step $t \in \{0, 1, ..., n\}$
\begin{align*}
    c^{t}_{k\rightarrow k', i} = 
    \begin{cases}
    \frac{n-t}{n}, & \text{if } i=k\\
    \frac{t}{n}, & \text{if } i=k'\\
    0, & \text{otherwise}
    \end{cases}
\end{align*}
Using this definition, we have for morphing from cervical to lumbar at step $t$
\begin{align*}
    c^t_{1\rightarrow 3} = \frac{1}{n}\begin{bmatrix}n-t\\0\\t\end{bmatrix}
\end{align*}
\subsection{Privacy}
To evaluate privacy, we created a list of candidate samples composed of 1000 images. 1/3 from training, 1/3 from validation (20\% of F2305 exclusively used in the privacy assessment) and 1/3 from test. For pairwise attacks, the comparison between synthetic and candidate is done in pixel space and embedding space. For the pixel space, similarity is computed as the minimum Euclidean distance between one candidate and all the synthetic samples. The embedding distance is obtained by training a UMAP trained on 5000 images (3000 from the train set, 1000 from the validation set, 1000 from the test set). The UMAP compresses the samples down to 64 features, the feature distance is obtained as the minimum Euclidean distance on features. For the distribution attacks, the threshold for 2 points to be considered as neighbours is that their distance is part of the 1st percentile in pixel space and 0.1st percentile in embedding space.

\section{Data and Code availability}
The real datasets (F2305, A2209) used for training and testing are part of an ongoing clinical trial and not publicly shareable. Instead, a synthetic version of the training dataset (with similar sample size and distribution) is shared as a substitute. Paper code and synthetic datasets can be found at \url{https://github.com/tcoroller/pGAN}. Novartis is committed to sharing with qualified external researchers’ access to patient-level data and supporting clinical documents from eligible studies. These requests are reviewed and approved based on scientific merit. All data provided are anonymized to respect the privacy of patients who have participated in the trial in line with applicable laws and regulations. The data may be requested from the corresponding author of the manuscript. The protocol would be made available on request by contacting the journal or the corresponding author.


\section{Acknowledgements}
The authors thank the study investigators for their contributions. The studies were funded by Novartis Pharma AG, Basel, Switzerland, in accordance with Good Publication Practice (GPP3) guidelines (\url{http://www.ismpp.org/gpp3}). 

{\small
\bibliographystyle{ieeetr}
\bibliography{egbib}

\begin{thebibliography}{10}

\bibitem{fauw}
J.~De~Fauw, J.~R. Ledsam, B.~Romera-Paredes, S.~Nikolov, N.~Tomasev,
  S.~Blackwell, H.~Askham, X.~Glorot, B.~O’Donoghue, D.~Visentin, {\em
  et~al.}, ``Clinically applicable deep learning for diagnosis and referral in
  retinal disease,'' {\em Nature medicine}, vol.~24, no.~9, pp.~1342--1350,
  2018.

\bibitem{schmauch}
B.~Schmauch, A.~Romagnoni, E.~Pronier, C.~Saillard, P.~Maill{\'e},
  J.~Calderaro, A.~Kamoun, M.~Sefta, S.~Toldo, M.~Zaslavskiy, {\em et~al.}, ``A
  deep learning model to predict rna-seq expression of tumours from whole slide
  images,'' {\em Nature communications}, vol.~11, no.~1, pp.~1--15, 2020.

\bibitem{hosny}
A.~Hosny, C.~Parmar, T.~P. Coroller, P.~Grossmann, R.~Zeleznik, A.~Kumar,
  J.~Bussink, R.~J. Gillies, R.~H. Mak, and H.~J. Aerts, ``Deep learning for
  lung cancer prognostication: a retrospective multi-cohort radiomics study,''
  {\em PLoS medicine}, vol.~15, no.~11, p.~e1002711, 2018.

\bibitem{xu}
Y.~Xu, A.~Hosny, R.~Zeleznik, C.~Parmar, T.~Coroller, I.~Franco, R.~H. Mak, and
  H.~J. Aerts, ``Deep learning predicts lung cancer treatment response from
  serial medical imaging,'' {\em Clinical Cancer Research}, vol.~25, no.~11,
  pp.~3266--3275, 2019.

\bibitem{pubmed}
``{PubMed}.'' \url{https://pubmed.ncbi.nlm.nih.gov}.

\bibitem{naos}
{National Academies of Sciences, Engineering and Medicine and others}, {\em
  Reproducibility and replicability in science}.
\newblock National Academies Press, 2019.

\bibitem{smidt}
H.~J. Smidt and O.~Jokonya, ``The challenge of privacy and security when using
  technology to track people in times of covid-19 pandemic,'' {\em Procedia
  Computer Science}, vol.~181, pp.~1018--1026, 2021.

\bibitem{tom}
E.~Tom, P.~A. Keane, M.~Blazes, L.~R. Pasquale, M.~F. Chiang, A.~Y. Lee, and
  C.~S. Lee, ``Protecting data privacy in the age of ai-enabled
  ophthalmology,'' {\em Translational Vision Science \& Technology}, vol.~9,
  no.~2, pp.~36--36, 2020.

\bibitem{wing}
J.~M. Wing, ``Ten research challenge areas in data science,'' {\em arXiv
  preprint arXiv:2002.05658}, 2020.

\bibitem{cannataci}
J.~Cannataci, V.~Falce, and O.~Pollicino, {\em Legal Challenges of Big Data}.
\newblock Edward Elgar Publishing, 2020.

\bibitem{grama}
J.~L. Grama, {\em Legal and Privacy Issues in Information Security}.
\newblock Jones \& Bartlett Learning, 2020.

\bibitem{prior}
F.~Prior, K.~Smith, A.~Sharma, J.~Kirby, L.~Tarbox, K.~Clark, W.~Bennett,
  T.~Nolan, and J.~Freymann, ``The public cancer radiology imaging collections
  of the cancer imaging archive,'' {\em Scientific data}, vol.~4, no.~1,
  pp.~1--7, 2017.

\bibitem{emam}
K.~El~Emam and L.~Arbuckle, {\em Anonymizing health data: case studies and
  methods to get you started}.
\newblock " O'Reilly Media, Inc.", 2013.

\bibitem{dwork}
C.~Dwork, ``Differential privacy: A survey of results,'' in {\em International
  conference on theory and applications of models of computation}, pp.~1--19,
  Springer, 2008.

\bibitem{fan}
L.~Fan, ``Image pixelization with differential privacy,'' in {\em Data and
  Applications Security and Privacy XXXII} (F.~Kerschbaum and S.~Paraboschi,
  eds.), (Cham), pp.~148--162, Springer International Publishing, 2018.

\bibitem{abadi}
M.~Abadi, A.~Chu, I.~Goodfellow, H.~B. McMahan, I.~Mironov, K.~Talwar, and
  L.~Zhang, ``Deep learning with differential privacy,'' in {\em Proceedings of
  the 2016 ACM SIGSAC Conference on Computer and Communications Security}, CCS
  '16, (New York, NY, USA), p.~308–318, Association for Computing Machinery,
  2016.

\bibitem{ryffel}
T.~Ryffel, A.~Trask, M.~Dahl, B.~Wagner, J.~Mancuso, D.~Rueckert, and
  J.~Passerat-Palmbach, ``A generic framework for privacy preserving deep
  learning,'' {\em arXiv preprint arXiv:1811.04017}, 2018.

\bibitem{goodfellow}
I.~J. Goodfellow, J.~Pouget-Abadie, M.~Mirza, B.~Xu, D.~Warde-Farley, S.~Ozair,
  A.~Courville, and Y.~Bengio, ``Generative adversarial networks,'' {\em arXiv
  preprint arXiv:1406.2661}, 2014.

\bibitem{karras}
T.~Karras, S.~Laine, and T.~Aila, ``A style-based generator architecture for
  generative adversarial networks,'' in {\em Proceedings of the IEEE/CVF
  Conference on Computer Vision and Pattern Recognition}, pp.~4401--4410, 2019.

\bibitem{mcinnes}
L.~McInnes, J.~Healy, and J.~Melville, ``Umap: Uniform manifold approximation
  and projection for dimension reduction,'' {\em arXiv preprint
  arXiv:1802.03426}, 2018.

\bibitem{vandermaaten}
L.~Van~der Maaten and G.~Hinton, ``Visualizing data using t-sne.,'' {\em
  Journal of machine learning research}, vol.~9, no.~11, 2008.

\bibitem{he}
K.~He, X.~Zhang, S.~Ren, and J.~Sun, ``Deep residual learning for image
  recognition,'' in {\em Proceedings of the IEEE conference on computer vision
  and pattern recognition}, pp.~770--778, 2016.

\bibitem{kazeminia}
S.~Kazeminia, C.~Baur, A.~Kuijper, B.~van Ginneken, N.~Navab, S.~Albarqouni,
  and A.~Mukhopadhyay, ``Gans for medical image analysis,'' {\em Artificial
  Intelligence in Medicine}, p.~101938, 2020.

\bibitem{yi}
X.~Yi, E.~Walia, and P.~Babyn, ``Generative adversarial network in medical
  imaging: A review,'' {\em Medical image analysis}, vol.~58, p.~101552, 2019.

\bibitem{chen}
D.~Chen, N.~Yu, Y.~Zhang, and M.~Fritz, ``Gan-leaks: A taxonomy of membership
  inference attacks against generative models,'' in {\em Proceedings of the
  2020 ACM SIGSAC Conference on Computer and Communications Security},
  pp.~343--362, 2020.

\bibitem{shin}
H.-C. Shin, N.~A. Tenenholtz, J.~K. Rogers, C.~G. Schwarz, M.~L. Senjem, J.~L.
  Gunter, K.~P. Andriole, and M.~Michalski, ``Medical image synthesis for data
  augmentation and anonymization using generative adversarial networks,'' in
  {\em International workshop on simulation and synthesis in medical imaging},
  pp.~1--11, Springer, 2018.

\bibitem{han}
C.~Han, H.~Hayashi, L.~Rundo, R.~Araki, W.~Shimoda, S.~Muramatsu, Y.~Furukawa,
  G.~Mauri, and H.~Nakayama, ``Gan-based synthetic brain mr image generation,''
  in {\em 2018 IEEE 15th International Symposium on Biomedical Imaging (ISBI
  2018)}, pp.~734--738, IEEE, 2018.

\bibitem{fossen-romsaas}
S.~Fossen-Romsaas and A.~Storm-Johannessen, ``Synthesizing skin lesion images
  using generative adversarial networks,'' Master's thesis, The University of
  Bergen, 2020.

\bibitem{frid-adar}
M.~Frid-Adar, I.~Diamant, E.~Klang, M.~Amitai, J.~Goldberger, and H.~Greenspan,
  ``Gan-based synthetic medical image augmentation for increased cnn
  performance in liver lesion classification,'' {\em Neurocomputing}, vol.~321,
  pp.~321--331, 2018.

\bibitem{zhang}
H.~Zhang, Z.~Huang, and Z.~Lv, ``Medical image synthetic data augmentation
  using gan,'' in {\em Proceedings of the 4th International Conference on
  Computer Science and Application Engineering}, pp.~1--6, 2020.

\bibitem{mukherjee}
S.~Mukherjee, Y.~Xu, A.~Trivedi, and J.~L. Ferres, ``Protecting gans against
  privacy attacks by preventing overfitting,'' 2020.

\bibitem{jordon}
J.~Jordon, J.~Yoon, and M.~Van Der~Schaar, ``Pate-gan: Generating synthetic
  data with differential privacy guarantees,'' in {\em International Conference
  on Learning Representations}, 2018.

\bibitem{beaulieu}
B.~K. Beaulieu-Jones, Z.~S. Wu, C.~Williams, R.~Lee, S.~P. Bhavnani, J.~B.
  Byrd, and C.~S. Greene, ``Privacy-preserving generative deep neural networks
  support clinical data sharing,'' {\em Circulation: Cardiovascular Quality and
  Outcomes}, vol.~12, no.~7, p.~e005122, 2019.

\bibitem{wang}
Z.~Wang, A.~C. Bovik, H.~R. Sheikh, and E.~P. Simoncelli, ``Image quality
  assessment: from error visibility to structural similarity,'' {\em IEEE
  transactions on image processing}, vol.~13, no.~4, pp.~600--612, 2004.

\bibitem{baeten}
D.~Baeten, J.~Sieper, J.~Braun, X.~Baraliakos, M.~Dougados, P.~Emery,
  A.~Deodhar, B.~Porter, R.~Martin, M.~Andersson, {\em et~al.}, ``Secukinumab,
  an interleukin-17a inhibitor, in ankylosing spondylitis,'' {\em New England
  journal of medicine}, vol.~373, no.~26, pp.~2534--2548, 2015.

\bibitem{jamaludin}
A.~Jamaludin, T.~Kadir, and A.~Zisserman, ``Spinenet: automated classification
  and evidence visualization in spinal mris,'' {\em Medical image analysis},
  vol.~41, pp.~63--73, 2017.

\bibitem{odena}
A.~Odena, C.~Olah, and J.~Shlens, ``Conditional image synthesis with auxiliary
  classifier gans,'' in {\em International conference on machine learning},
  pp.~2642--2651, PMLR, 2017.

\bibitem{kingma}
D.~P. Kingma and J.~Ba, ``Adam: A method for stochastic optimization,'' {\em
  arXiv preprint arXiv:1412.6980}, 2014.

\bibitem{efron}
B.~Efron, ``Better bootstrap confidence intervals,'' {\em Journal of the
  American statistical Association}, vol.~82, no.~397, pp.~171--185, 1987.

\end{thebibliography}
}

\beginsupplement

\begin{figure*}
\begin{center}
\includegraphics[width=\textwidth]{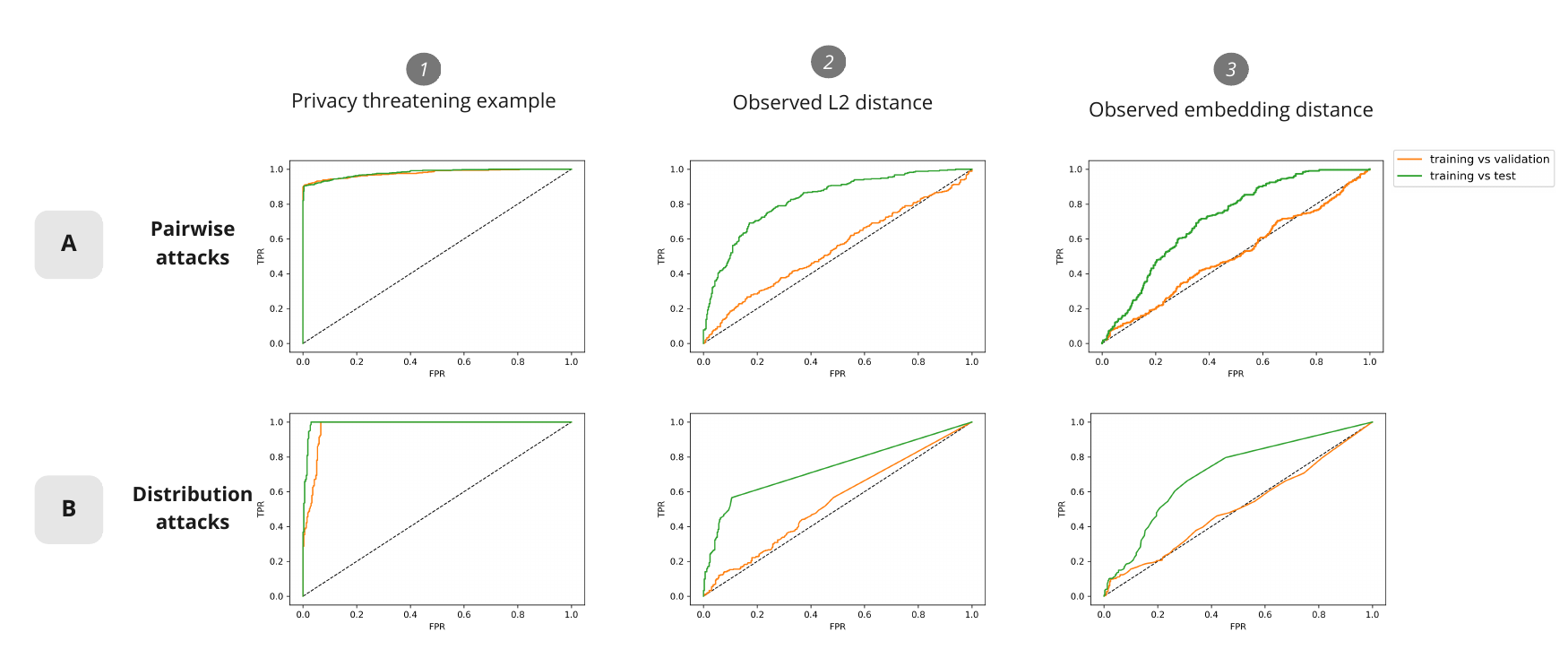}
\end{center}
   \caption{\textbf{ROC curves.} ROC for classification of candidate samples from train vs validation (orange) and train vs test (green). A) Pairwise attacks: The classification is under the assumption candidates with the lowest distances are likely from training. B) Distribution attacks: The classification is done by associating large clusters to training samples. A high AUC means it is easy to classify training from other samples.}
\label{fig:s03}
\end{figure*}

\begin{figure*}
\begin{center}
\includegraphics[width=\textwidth]{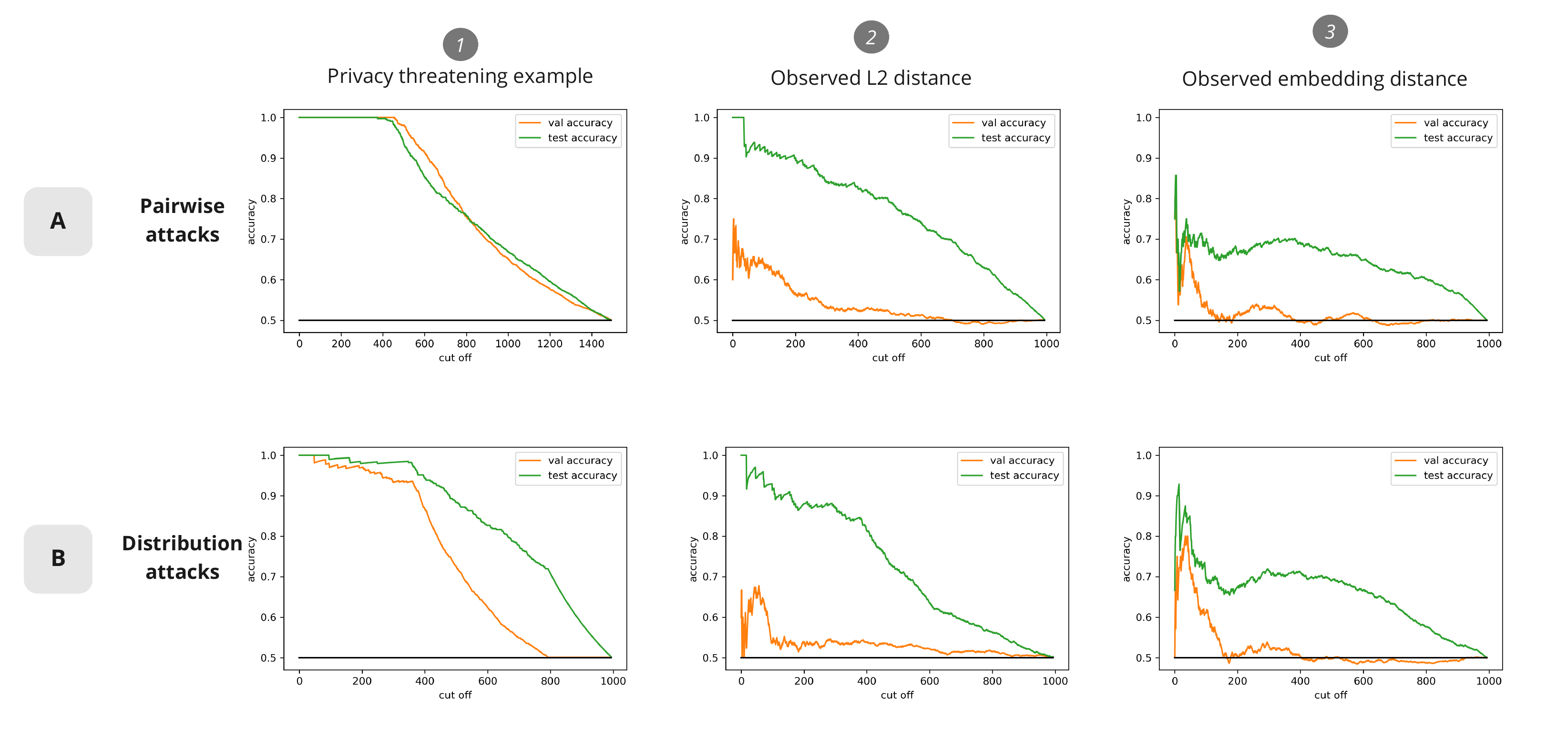}
\end{center}
   \caption{\textbf{Cut-off curves.} Cut off defines the threshold for a point to be considered an outlier (outlier means the image is probably from training). The curve orange shows the proportion of outliers from training vs outliers from validation. The green curve shows the proportion of outliers from training vs outliers from test. A privacy threatening case correspond to almost all of outliers coming from the training set.}
\label{fig:s04}
\end{figure*}

\begin{table*}[t]
\centering
    \begin{tabular}{|c|c|c|c|c|c|c|}
    \hline
    & \multicolumn{6}{|c|}{Pairwise attacks}\\ \cline{2-7}
    &\multicolumn{2}{|c|}{Toy example}& \multicolumn{2}{|c|}{L2 distance} & \multicolumn{2}{|c|}{Embedding distance}\\ \hline
    \textbf{Cut off} & \textbf{50}& \textbf{333}& \textbf{50}& \textbf{333}& \textbf{50}& \textbf{333} \\ \hline
    Train proportion & 1.00 & 0.89 & 0.58 & 0.48 & 0.51 & 0.43\\ \hline
    Val proportion & 0.00 & 0.04 & 0.36 & 0.43 & 0.27 & 0.38\\ \hline
    Test proportion & 0.00 & 0.07 & 0.05 & 0.09 & 0.22 & 0.19\\ \hline
    \end{tabular}
        \begin{tabular}{|c|c|c|c|c|c|c|}
    \hline
    & \multicolumn{6}{|c|}{Distribution attacks}\\ \cline{2-7}
    &\multicolumn{2}{|c|}{Toy example}& \multicolumn{2}{|c|}{L2 distance} & \multicolumn{2}{|c|}{Embedding distance}\\ \hline
    \textbf{Cut off} & \textbf{50}& \textbf{333}& \textbf{50}& \textbf{333}& \textbf{50}& \textbf{333} \\ \hline
    Train proportion & 0.93 & 0.91 & 0.64 & 0.49 & 0.62 & 0.43\\ \hline
    Val proportion & 0.05 & 0.07 & 0.33 & 0.42 & 0.24 & 0.39\\ \hline
    Test proportion & 0.02 & 0.02 & 0.04 & 0.09 & 0.15 & 0.18\\ \hline
    \end{tabular}
\caption{\textbf{Classification of candidate origin.} Proportion of origin for different candidates at given cut-off. Cut off of 50 means the top 50 largest clusters (pairwise attacks) or smallest distances (distribution attacks) are considered outliers (50 is top 5\% of the 1000 candidate images). In extreme privacy threatening scenarios, we expect that all training sample can be identified as outliers meaning they would represent 100\% of the top 333 candidates.}
\label{tab:label}
\end{table*}
\end{document}